\documentclass[letterpaper]{article} 
\usepackage{aaai2026}  
\usepackage{times}  
\usepackage{helvet}  
\usepackage{courier}  
\usepackage[hyphens]{url}  
\usepackage{graphicx} 
\usepackage{natbib}  
\usepackage{caption} 
\usepackage{algorithm}
\usepackage{algorithmic}

\usepackage{svg}
\usepackage{graphicx}
\usepackage{amsmath}
\usepackage{xspace}
\usepackage{multirow}
\usepackage{amssymb}
\usepackage{bbold}
\usepackage{booktabs}
\usepackage{pgfplots}
\usepackage{adjustbox}

\usepackage{colortbl}

\usepackage{tikz}
\usetikzlibrary{
    arrows,
    automata,
    backgrounds,
    calc,
    decorations.shapes,
    fit,
    mindmap,
    patterns,
    petri,
    positioning,
    shadows,
    shapes,
    shapes.gates.logic.US,
    trees
}
\definecolor{tol-hc-red}{HTML}{BB5566} 
\definecolor{tol-hc-blue}{HTML}{004488} 
\definecolor{tol-hc-yellow}{HTML}{DDAA33} 
\definecolor{tol-legend}{HTML}{666666}

\usepackage{newfloat}
\usepackage{listings}
\DeclareCaptionStyle{ruled}{labelfont=normalfont,labelsep=colon,strut=off} 
\floatstyle{ruled}
\newfloat{listing}{tb}{lst}{}
\floatname{listing}{Listing}
\title{SL-CBM: Enhancing Concept Bottleneck Models with Semantic Locality for Better Interpretability}
\author {
    Hanwei Zhang \textsuperscript{\rm 1},
    Luo Cheng \textsuperscript{\rm 2,\rm 5},
    Rui Wen \textsuperscript{\rm 3},
    Yang Zhang \textsuperscript{\rm 4},
    Lijun Zhang \textsuperscript{\rm 5},
    Holger Hermanns \textsuperscript{\rm 1}
}
\affiliations {
    \textsuperscript{\rm 1} Saarland University\\
    \textsuperscript{\rm 2} University of Chinese Academy of Sciences\\
    \textsuperscript{\rm 3} Institute of Science Tokyo\\
    \textsuperscript{\rm 4} CISPA Helmholtz Center for Information Security\\
    \textsuperscript{\rm 5} Key Laboratory of System Software (Chinese Academy of Sciences) and State Key Laboratory of Computer Science, Institute of Software, Chinese Academy of Science\\
    \{zhang,hermanns\}@depend.uni-saarland.de, \{chengluo,zhanglj\}@ios.ac.cn, rui.w.b9ce@m.isct.ac.jp, zhang@cispa.de
}

\usepackage{bibentry}

\usepackage{tikz}
\usetikzlibrary{arrows,shapes,calc,matrix,fit,backgrounds}
\usepackage{pgfplots}
\usepackage{pgfplotstable}
\pgfplotsset{compat=1.9}

\usepackage{xstring}

\usepgfplotslibrary{external}

\IfBeginWith*{\jobname}{fig/extern/}{\finalcopy}{}

\tikzset{every mark/.append style={solid}}
\pgfplotsset{
	grid=both, width=\columnwidth, try min ticks=5,
	every axis/.append style={font=\scriptsize},
	every axis plot/.append style={thick,mark=none,mark size=1.2,tension=0.18},
	legend cell align=left, legend style={fill opacity=0.8},
}

\pgfplotsset{
	dash/.style={mark=o,dashed,opacity=0.7},
	dott/.style={mark=o,dotted,opacity=0.7},
}

\usepgfplotslibrary{fillbetween}

\begin{document}

\maketitle

\begin{abstract}

Explainable AI (XAI) is crucial for building transparent and trustworthy machine learning systems, especially in high-stakes domains. Concept Bottleneck Models (CBMs) have emerged as a promising ante-hoc approach that provides interpretable, concept-level explanations by explicitly modeling human-understandable concepts. However, existing CBMs often suffer from poor locality faithfulness, failing to spatially align concepts with meaningful image regions, which limits their interpretability and reliability. 
In this work, we propose SL-CBM (CBM with Semantic Locality), a novel extension that enforces locality faithfulness by generating spatially coherent saliency maps at both concept and class levels. SL-CBM integrates a $1 \times 1$ convolutional layer with a cross-attention mechanism to enhance alignment between concepts, image regions, and final predictions. Unlike prior methods, SL-CBM produces faithful saliency maps inherently tied to the model’s internal reasoning, facilitating more effective debugging and intervention. Extensive experiments on image datasets demonstrate that SL-CBM substantially improves locality faithfulness, explanation quality, and intervention efficacy while maintaining competitive classification accuracy. Our ablation studies highlight the importance of contrastive and entropy-based regularization for balancing accuracy, sparsity, and faithfulness. Overall, SL-CBM bridges the gap between concept-based reasoning and spatial explainability, setting a new standard for interpretable and trustworthy concept-based models. 

\end{abstract}

%
\begin{links}
\small
\link{Code}{https://github.com/Uzukidd/sl-cbm}
\link{RIVAL10}{https://mmoayeri.github.io/RIVAL10/index.html}
\link{CUB}{ https://www.vision.caltech.edu/datasets/cub_200_2011/}
\link{PCBM}{https://github.com/mertyg/post-hoc-cbm}
\link{CCS}{https://github.com/NMS05/Improving-Concept-Alignment-in-Vision-Language-Concept-Bottleneck-Models}\link{CLIP-ViT}{https://huggingface.co/laion/CLIP-ViT-B-16-laion2B-s34B-b88K}

\end{links}

\newcommand{\head}[1]{{\smallskip\noindent\textbf{#1}}}
\newcommand{\alert}[1]{{\color{red}{#1}}}
\newcommand{\sm}{\scriptsize}
\newcommand{\eq}[1]{(\ref{eq:#1})}

\newcommand{\Th}[1]{\textsc{#1}}
\newcommand{\mr}[2]{\multirow{#1}{*}{#2}}
\newcommand{\mc}[2]{\multicolumn{#1}{c}{#2}}
\newcommand{\tb}[1]{\textbf{#1}}
\newcommand{\ch}{\checkmark}

\newcommand{\blue}[1]{{\textcolor{blue}{#1}}}
\newcommand{\green}[1]{{\textcolor{green}{#1}}}
\newcommand{\gray}[1]{{\textcolor{gray}{#1}}}

\newcommand{\citeme}[1]{\red{[XX]}}
\newcommand{\refme}[1]{\red{(XX)}}

\newcommand{\fig}[2][1]{\includegraphics[width=#1\linewidth]{fig/#2}}
\newcommand{\figh}[2][1]{\includegraphics[height=#1\linewidth]{fig/#2}}
\newcommand{\figa}[2][1]{\includegraphics[width=#1]{fig/#2}}
\newcommand{\figah}[2][1]{\includegraphics[height=#1]{fig/#2}}

\newcommand{\tran}{^\top}
\newcommand{\mtran}{^{-\top}}
\newcommand{\zcol}{\mathbf{0}}
\newcommand{\zrow}{\zcol\tran}

\newcommand{\ind}{\mathbbm{1}}
\newcommand{\expect}{\mathbb{E}}
\newcommand{\nat}{\mathbb{N}}
\newcommand{\zahl}{\mathbb{Z}}
\newcommand{\real}{\mathbb{R}}
\newcommand{\proj}{\mathbb{P}}
\newcommand{\prob}{\operatorname{P}}
\newcommand{\normal}{\mathcal{N}}

\newcommand{\mif}{\textrm{if}\ }
\newcommand{\other}{\textrm{otherwise}}
\newcommand{\minimize}{\textrm{minimize}\ }
\newcommand{\maximize}{\textrm{maximize}\ }
\newcommand{\st}{\textrm{subject\ to}\ }

\newcommand{\id}{\operatorname{id}}
\newcommand{\const}{\operatorname{const}}
\newcommand{\sgn}{\operatorname{sgn}}
\newcommand{\var}{\operatorname{Var}}
\newcommand{\mean}{\operatorname{mean}}
\newcommand{\trace}{\operatorname{tr}}
\newcommand{\diag}{\operatorname{diag}}
\newcommand{\vect}{\operatorname{vec}}
\newcommand{\cov}{\operatorname{cov}}
\newcommand{\sign}{\operatorname{sign}}
\newcommand{\prj}{\operatorname{proj}}

\newcommand{\softmax}{\operatorname{softmax}}
\newcommand{\clip}{\operatorname{clip}}

\newcommand{\defn}{\mathrel{:=}}
\newcommand{\peq}{\mathrel{+\!=}}
\newcommand{\meq}{\mathrel{-\!=}}

\newcommand{\paren}[1]{\left({#1}\right)}
\newcommand{\mat}[1]{\left[{#1}\right]}
\newcommand{\set}[1]{\left\{{#1}\right\}}
\newcommand{\floor}[1]{\left\lfloor{#1}\right\rfloor}
\newcommand{\ceil}[1]{\left\lceil{#1}\right\rceil}
\newcommand{\inner}[1]{\left\langle{#1}\right\rangle}
\newcommand{\norm}[1]{\left\|{#1}\right\|}
\newcommand{\abs}[1]{\left|{#1}\right|}
\newcommand{\frob}[1]{\norm{#1}_F}
\newcommand{\card}[1]{\left|{#1}\right|\xspace}

\newcommand{\diff}{\mathrm{d}}
\newcommand{\der}[3][]{\frac{\diff^{#1}#2}{\diff#3^{#1}}}
\newcommand{\ider}[3][]{\diff^{#1}#2/\diff#3^{#1}}
\newcommand{\pder}[3][]{\frac{\partial^{#1}{#2}}{\partial{{#3}^{#1}}}}
\newcommand{\ipder}[3][]{\partial^{#1}{#2}/\partial{#3^{#1}}}
\newcommand{\dder}[3]{\frac{\partial^2{#1}}{\partial{#2}\partial{#3}}}

\newcommand{\wb}[1]{\overline{#1}}
\newcommand{\wt}[1]{\widetilde{#1}}

\def\xssp{\hspace{1pt}}
\def\ssp{\hspace{3pt}}
\def\msp{\hspace{5pt}}
\def\lsp{\hspace{12pt}}

\newcommand{\cA}{\mathcal{A}}
\newcommand{\cB}{\mathcal{B}}
\newcommand{\cC}{\mathcal{C}}
\newcommand{\cD}{\mathcal{D}}
\newcommand{\cE}{\mathcal{E}}
\newcommand{\cF}{\mathcal{F}}
\newcommand{\cG}{\mathcal{G}}
\newcommand{\cH}{\mathcal{H}}
\newcommand{\cI}{\mathcal{I}}
\newcommand{\cJ}{\mathcal{J}}
\newcommand{\cK}{\mathcal{K}}
\newcommand{\cL}{\mathcal{L}}
\newcommand{\cM}{\mathcal{M}}
\newcommand{\cN}{\mathcal{N}}
\newcommand{\cO}{\mathcal{O}}
\newcommand{\cP}{\mathcal{P}}
\newcommand{\cQ}{\mathcal{Q}}
\newcommand{\cR}{\mathcal{R}}
\newcommand{\cS}{\mathcal{S}}
\newcommand{\cT}{\mathcal{T}}
\newcommand{\cU}{\mathcal{U}}
\newcommand{\cV}{\mathcal{V}}
\newcommand{\cW}{\mathcal{W}}
\newcommand{\cX}{\mathcal{X}}
\newcommand{\cY}{\mathcal{Y}}
\newcommand{\cZ}{\mathcal{Z}}

\newcommand{\vA}{\mathbf{A}}
\newcommand{\vB}{\mathbf{B}}
\newcommand{\vC}{\mathbf{C}}
\newcommand{\vD}{\mathbf{D}}
\newcommand{\vE}{\mathbf{E}}
\newcommand{\vF}{\mathbf{F}}
\newcommand{\vG}{\mathbf{G}}
\newcommand{\vH}{\mathbf{H}}
\newcommand{\vI}{\mathbf{I}}
\newcommand{\vJ}{\mathbf{J}}
\newcommand{\vK}{\mathbf{K}}
\newcommand{\vL}{\mathbf{L}}
\newcommand{\vM}{\mathbf{M}}
\newcommand{\vN}{\mathbf{N}}
\newcommand{\vO}{\mathbf{O}}
\newcommand{\vP}{\mathbf{P}}
\newcommand{\vQ}{\mathbf{Q}}
\newcommand{\vR}{\mathbf{R}}
\newcommand{\vS}{\mathbf{S}}
\newcommand{\vT}{\mathbf{T}}
\newcommand{\vU}{\mathbf{U}}
\newcommand{\vV}{\mathbf{V}}
\newcommand{\vW}{\mathbf{W}}
\newcommand{\vX}{\mathbf{X}}
\newcommand{\vY}{\mathbf{Y}}
\newcommand{\vZ}{\mathbf{Z}}

\newcommand{\va}{\mathbf{a}}
\newcommand{\vb}{\mathbf{b}}
\newcommand{\vc}{\mathbf{c}}
\newcommand{\vd}{\mathbf{d}}
\newcommand{\ve}{\mathbf{e}}
\newcommand{\vf}{\mathbf{f}}
\newcommand{\vg}{\mathbf{g}}
\newcommand{\vh}{\mathbf{h}}
\newcommand{\vi}{\mathbf{i}}
\newcommand{\vj}{\mathbf{j}}
\newcommand{\vk}{\mathbf{k}}
\newcommand{\vl}{\mathbf{l}}
\newcommand{\vm}{\mathbf{m}}
\newcommand{\vn}{\mathbf{n}}
\newcommand{\vo}{\mathbf{o}}
\newcommand{\vp}{\mathbf{p}}
\newcommand{\vq}{\mathbf{q}}
\newcommand{\vr}{\mathbf{r}}
\newcommand{\Vs}{\mathbf{s}}
\newcommand{\vt}{\mathbf{t}}
\newcommand{\vu}{\mathbf{u}}
\newcommand{\vv}{\mathbf{v}}
\newcommand{\vw}{\mathbf{w}}
\newcommand{\vx}{\mathbf{x}}
\newcommand{\vy}{\mathbf{y}}
\newcommand{\vz}{\mathbf{z}}

\newcommand{\vone}{\mathbf{1}}
\newcommand{\vzero}{\mathbf{0}}

\newcommand{\valpha}{{\boldsymbol{\alpha}}}
\newcommand{\vbeta}{{\boldsymbol{\beta}}}
\newcommand{\vgamma}{{\boldsymbol{\gamma}}}
\newcommand{\vdelta}{{\boldsymbol{\delta}}}
\newcommand{\vepsilon}{{\boldsymbol{\epsilon}}}
\newcommand{\vzeta}{{\boldsymbol{\zeta}}}
\newcommand{\veta}{{\boldsymbol{\eta}}}
\newcommand{\vtheta}{{\boldsymbol{\theta}}}
\newcommand{\viota}{{\boldsymbol{\iota}}}
\newcommand{\vkappa}{{\boldsymbol{\kappa}}}
\newcommand{\vlambda}{{\boldsymbol{\lambda}}}
\newcommand{\vmu}{{\boldsymbol{\mu}}}
\newcommand{\vnu}{{\boldsymbol{\nu}}}
\newcommand{\vxi}{{\boldsymbol{\xi}}}
\newcommand{\vomikron}{{\boldsymbol{\omikron}}}
\newcommand{\vpi}{{\boldsymbol{\pi}}}
\newcommand{\vrho}{{\boldsymbol{\rho}}}
\newcommand{\vsigma}{{\boldsymbol{\sigma}}}
\newcommand{\vtau}{{\boldsymbol{\tau}}}
\newcommand{\vupsilon}{{\boldsymbol{\upsilon}}}
\newcommand{\vphi}{{\boldsymbol{\phi}}}
\newcommand{\vchi}{{\boldsymbol{\chi}}}
\newcommand{\vpsi}{{\boldsymbol{\psi}}}
\newcommand{\vomega}{{\boldsymbol{\omega}}}

\newcommand{\rLambda}{\mathrm{\Lambda}}
\newcommand{\rSigma}{\mathrm{\Sigma}}

\newcommand{\vLambda}{\bm{\rLambda}}
\newcommand{\vSigma}{\bm{\rSigma}}

\makeatletter
\newcommand*\bdot{\mathpalette\bdot@{.7}}
\newcommand*\bdot@[2]{\mathbin{\vcenter{\hbox{\scalebox{#2}{$\m@th#1\bullet$}}}}}
\makeatother

\makeatletter
\DeclareRobustCommand\onedot{\futurelet\@let@token\@onedot}
\def\@onedot{\ifx\@let@token.\else.\null\fi\xspace}

\def\eg{\emph{e.g}\onedot} \def\Eg{\emph{E.g}\onedot}
\def\ie{\emph{i.e}\onedot} \def\Ie{\emph{I.e}\onedot}
\def\cf{\emph{cf}\onedot} \def\Cf{\emph{Cf}\onedot}
\def\etc{\emph{etc}\onedot} \def\vs{\emph{vs}\onedot}
\def\wrt{w.r.t\onedot} \def\dof{d.o.f\onedot} \def\aka{a.k.a\onedot}
\def\etal{\emph{et al}\onedot}
\makeatother

\newcommand{\interventionPlot}[3]{%

    \pgfplotstableread{#1_intervention.txt}\datatable

    \addplot[name path=#1_mean, thick, #2] table[
        x=x,
        y=mean
    ] {\datatable};

    \addplot[name path=#1_upper, draw=none] table[
        x=x,
        y expr=\thisrow{mean} + \thisrow{std}
    ] {\datatable};

    \addplot[name path=#1_lower, draw=none] table[
        x=x,
        y expr=\thisrow{mean} - \thisrow{std}
    ] {\datatable};

    \addplot[#2, fill opacity=0.10] fill between[
        of=#1_upper and #1_lower,
    ];
}

\section{Introduction}

\begin{figure}[htpb]
    \centering
    \setlength{\tabcolsep}{0.5pt}
    \renewcommand{\arraystretch}{0.5}
    \begin{tabular}{c c c c}
    \rotatebox{90}{Concept (Horn)} & 
    \includegraphics[width=0.3\linewidth]{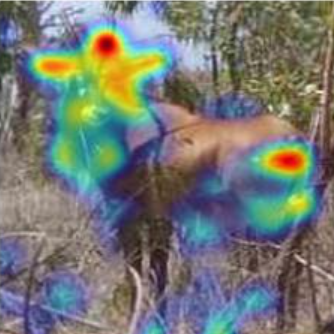}&    
    \includegraphics[width=0.3\linewidth]{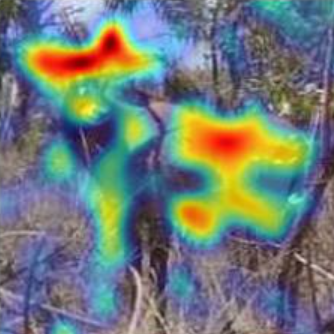}& 
    \includegraphics[width=0.3\linewidth]{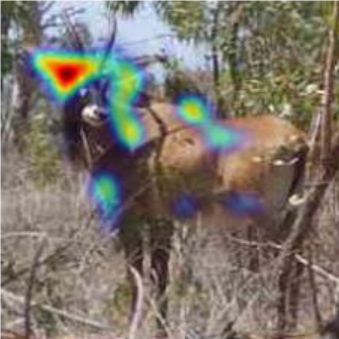}\\
    \rotatebox{90}{Class (Plane)}  &
    \includegraphics[trim={0mm 0mm 0mm 1.7mm},clip, width=0.3\linewidth]{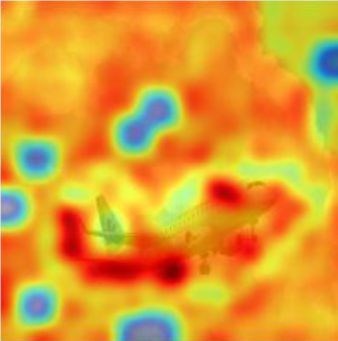}&   
    \includegraphics[trim={0mm 0mm 0mm 1.5mm},clip, width=0.3\linewidth]{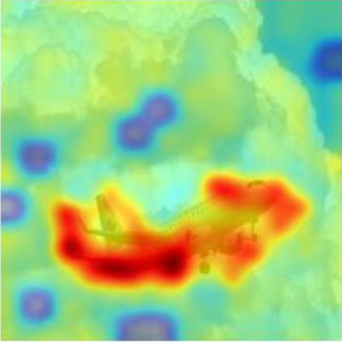}&
    \includegraphics[width=0.3\linewidth]{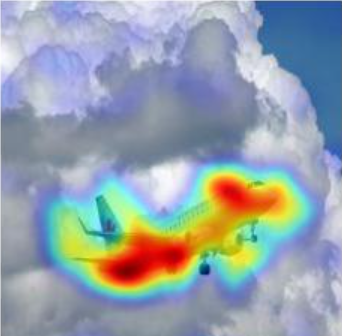}\\
    & PCBM & CSS & SL-CBM (ours)
    \end{tabular}    
    \caption{Saliency maps of state-of-the-art CBMs and SL-CBM at both concept and class levels. Saliency maps of PCBM~\cite{yuksekgonul2022post} and CSS~\cite{selvaraj2024improving} are generated using GradCAM~\cite{selvaraju2017grad}, while SL-CBM produces its own saliency maps.}
    \label{fig:teaser}
\end{figure}

The widespread adoption of AI has heightened concerns around AI alignment, with explainable AI (XAI) increasingly recognized as central to ensuring aligned, transparent, and trustworthy systems. Among existing XAI techniques, saliency methods have gained popularity due to their intuitive, low-cost, and post-hoc nature. However, studies have shown that these methods often lack faithfulness and fail to deliver reliable explanations in high-stakes settings~\cite{zhang2024saliency,kares2025makes}.
As a more faithful alternative, Concept Bottleneck Models (CBMs)~\cite{koh2020concept} incorporate a layer of human-interpretable concepts between inputs and predictions, providing concept-level insights into model behavior. Unlike post-hoc methods, CBMs operate in an ante-hoc manner, ensuring that explanations directly reflect the model’s decision-making process. They also support human intervention through concept correction~\cite{chauhan2023interactive}. Due to their practicality, low conversion cost~\cite{yuksekgonul2022post}, and recent extensions to multimodal settings~\cite{selvaraj2024improving}, CBMs are gaining increasing attention as a viable framework for explainable and aligned AI.

However, existing CBMs struggle with locality faithfulness, particularly in image-based tasks where concepts often fail to align with relevant image regions or contribute meaningfully to the final prediction~\cite{margeloiu2021concept, raman2023concept, furby2023towards}. 
Improving locality faithfulness is key to making CBM explanations more reliable and understandable, especially in high-risk settings, where it enables better human-guided interventions, though current CBM interventions mainly rely on automation, limiting human control.
Efforts to improve concept trustworthiness have explored aligning concepts with classes and enforcing cross-layer/image alignment~\cite{huang2024concept, selvaraj2024improving}, yet existing approaches remain limited in capturing meaningful alignment between concepts and images. To the best of our knowledge, our work is the first to directly address these limitations by explicitly enhancing the locality faithfulness of CBMs.
As illustrated in Figure~\ref{fig:teaser}, existing methods such as PCBM~\cite{yuksekgonul2022post} and CSS~\cite{selvaraj2024improving} often highlight irrelevant regions, \eg, in generating saliency maps for the concept \emph{Horn}, or even more drastically, for the class \emph{Plane}.

Inspired by CAM~\cite{zhou2016learning} and CBMs~\cite{koh2020concept}, we propose \emph{CBM with Semantic Locality (SL-CBM)}, a simple yet effective structure to enforce locality faithfulness by aligning concept saliency maps with images' concept projection. SL-CBM generates both concept- and class-level saliency maps alongside predictions, with class saliency maps derived by linearly combining concept maps using class-specific weights. As shown in Figure \ref{fig:teaser}, SL-CBM better localizes concepts like \emph{Horn} and focuses on objects like \emph{Plane} without highlighting irrelevant areas. Quantitative evaluation with XAI and localization metrics confirms SL-CBM’s improved locality faithfulness, explainability, and intervention.
{In summary, our contributions are as follows:}
\begin{itemize}
    \item {We propose SL-CBM, a model that generates saliency maps and concept-based explanations to enhance human understanding of the decision-making process.}
    \item {By enforcing locality faithfulness, SL-CBM enhances the alignment between image space, concept space, and class predictions, thereby improving the model’s interpretability and reliability.}
    \item {We systematically evaluate SL-CBM’s concept- and class-level accuracy, locality faithfulness, and intervention effectiveness, demonstrating improved explanation accuracy and faithfulness with potential to enhance human understanding and model refinement.} 
\end{itemize}
\section{Related Work}

\paragraph{Concept Bottleneck Models.}
The use of concept bottlenecks in deep neural networks for task-specific solutions or explainability is well-established~\cite{yi2018neural,chen2020concept,losch2019interpretability,kim2018interpretability,de2024visual}. However, \citet{koh2020concept} first formally defined CBMs as a concept project backbone network paired with a classifier. CBMs address three goals: \emph{Interpretability} (identifying important concepts), \emph{Predictability} (predicting targets from concepts), and \emph{Intervenability} (improving predictions by replacing concept values with ground truth). Due to interpretability, intervenability, and adaptability~\cite{dominici2024anycbms}, CBMs have gained prominence in XAI.

Recent research focuses on enhancing CBMs through improved concept quality and broader applicability. One direction refines concept annotations using language models: \eg, GPT-3-generated concept sets \cite{oikarinen2023label} and LaBo’s submodular selection of discriminative, CLIP-aligned concepts \cite{yang2023language}. Another approach replaces rigid concept definitions with \emph{soft concepts}: PCBMs enable data-efficient conversion of pretrained models into CBMs \cite{yuksekgonul2022post}, while ProbCBM introduces probabilistic embeddings to handle data ambiguity \cite{kim2023probabilistic}. Further innovations include autoregressive concept predictors~\cite{havasi2022addressing} and ChatGPT-guided concept augmentation~\cite{tan2024interpreting}.
Our research does not focus on concept quality; therefore, we use a predefined concept set across all CBM models for fair comparison. While performance may be improved with a refined concept set, this lies beyond the scope of our study.

\paragraph{Locality Faithfulness of CBMs.}
Despite progress, CBMs face persistent issues in \emph{locality faithfulness}—ensuring concepts align with spatially or semantically localized input features. Studies reveal that CBMs often fail to learn localized concept representations: \citet{margeloiu2021concept} find limited concept interpretability using Integrated Gradients~\cite{sundararajan2017axiomatic}, while \citet{raman2023concept} demonstrate poor spatial and semantic locality. \citet{furby2023towards} corroborate this via LRP~\cite{bach2015pixel}, showing concepts rarely map to distinct input regions. To address this, \citet{huang2024concept} propose GradCAM-based~\cite{selvaraju2017grad} evaluation of concept trustworthiness, and \citet{selvaraj2024improving} advocate for locality alignment between concepts and classes. Interactive approaches, such as human-in-the-loop concept labeling \cite{chauhan2023interactive}, aim to improve faithfulness by grounding concepts in human oversight. These efforts primarily aim to enhance concept quality for better explanations. However, the issue of locality faithfulness has been identified but remains unaddressed. Our work aims to bridge this gap in the field.


\section{Method}

\begin{figure*}[hbpt]
    \centering
    \includegraphics[width=0.9\linewidth]{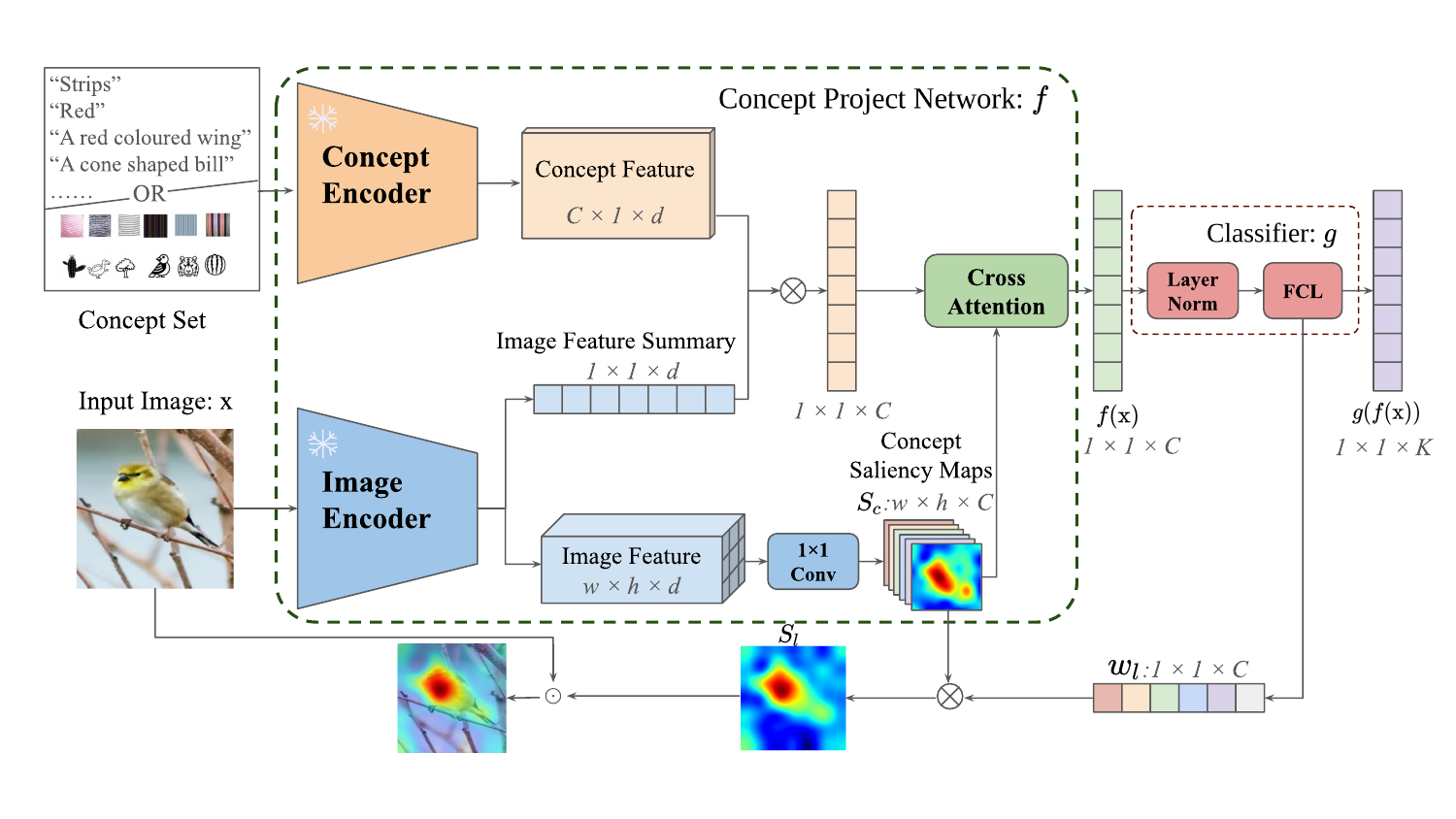}
    \caption{ SL-CBM Overview: Given an input image $\vx$, a concept set, the fixed concept and image encoders extract concept and image features, and an image feature summary. Projecting the image summary onto concept features yields a similarity vector. A $1 \times 1$ convolution generates concept saliency maps $S_\vc$, which, with the similarity vector, are refined via cross-attention into  $f(\vx)$. preserving locality and concept relevance. A classifier then produces logit $g(f(\vx))$, and class saliency map $S_l$ is computed by weighting $S_\vc$ with the class-specific FCL weight $\vw_l$.}
    \label{fig:framework}
\end{figure*}

\subsection{Problem Formulation}
Consider Concept Bottleneck Models (CBMs) as a pair $(f,g)$ consisting of a concept project network $f: \cX \to \real^{C}$, which maps an input image $\vx \in \cX$ to a concept space $\real^{C}$ containing $C$ predefined concepts, and a classifier $g: \real^C \to \real^K$, which maps the predicted concept embedding to one of $K$ target classes. 
Let $l_{gt}$ represent the ground truth class label of $\vx$ and let $\cC_{gt}$ denote the set of ground truth concept labels associated with this input. Let $k\defn |\cC_{gt}|$ be its cardinality. 
Define the predicted set $\cP$ as the concept indices of the top-$k$ values of $f(\vx)$:
\[
    \cP \defn \left\{ i\in \cI | f(\vx)_i \text{ is in the top-}k \text{ of the } \{f(\vx)_j\}_{j \in \cI} \right\},
\]
where $\cI$ is the full set of candidate concept indices and $f(\vx)_j$ denotes the value of $j^{th}$ concept $c^j$. Concept accuracy is evaluated based on the overlap between the predicted set $\cP$ and the ground truth set $\cC_{gt}$, \ie
\[
    \frac{|\cP \cap \cC_{gt}|}{k}.
\]
The class predicted label is given by 
\[l \defn \underset{i}{\arg \max}~g(f(\vx))_i,\] 
where $g(\cdot)_i$ denotes the predicted logit value of the class $i^{th}$. The CBM prediction is considered correct when $ l=l_{gt}$.

Let $S_{c^i}$ denote the saliency map for the $i^{th}$ concept $c^i$ and let $S_l$ denote the saliency map for the class $l$. 
The \emph{locality faithfulness} of CBMs can be defined at two levels: concept-level and class-level. At the concept level, the locality faithfulness ensures that the saliency map of the $i^{th}$ concept highlights the most relevant information for concept $c^i$. Thus when the input image is masked by the saliency map of concept $i$, it should emphasize the corresponding concept, formally expressed as \[\underset{j}{\arg \max}~f(\vx \odot S_{c^i})_j = i.\] At the class level, locality faithfulness aligns with traditional saliency maps, meaning that the saliency map should preserve the information relevant to the class prediction. This can be formulated as: \[\underset{i}{\arg \max}~g(f(\vx \odot S_{l_{gt}}))_i = \underset{i}{\arg \max}~g(f(\vx))_i=l_{gt}.\]
Our objective is to maximize accuracy while enhancing locality faithfulness at both the concept and class levels.

\subsection{CBM with Semantic Locality}

To improve CBM interpretability, we propose \emph{CBM with Semantic Locality (SL-CBM)}, which generates semantic saliency maps at both the concept and class levels. As Figure \ref{fig:framework} shows, SL-CBM extends the concept projection network by adding a branch to learn concept saliency, enabling localized explanations at the concept level. The concept feature is derived from a dedicated concept encoder spanning the concept subspace. While preserving the conventional projection of input embeddings, \ie image feature summary, onto the concept subspace, SL-CBM also derives a separate image feature for saliency generation from a shared image encoder. 
The concept-based image representation $f(x)$, used for classification, is generated by combining concept saliency maps and concept subspace projections via a cross-attention module, encouraging local, interpretable features.

\paragraph{Concept Project Network with Pre-trained Backbone.}
Existing CBMs operate in two main settings: (1) using a vision backbone with concept activation vectors (CAVs)~\cite{kim2018interpretability} to learn concept vectors as in PCBM~\cite{yuksekgonul2022post}, or (2) leveraging vision-language models for both image and concept embeddings. SL-CBM is compatible with any pretrained backbone, ensuring flexibility. With CNNs, it extracts image features from the last convolutional layer and uses average pooling for the image summary. With transformers, it uses spatial tokens as image features and the CLS token as the summary.

\paragraph{Saliency Maps at Concept and Class levels.}
To enable the concept projection network $f$ to generate saliency maps at both concept and class levels, we use a \emph{$1 \times 1$ convolution} to learn weights over spatial features, producing concept saliency maps $S_\vc$ that preserve semantic locality. To reinforce the faithfulness of concept saliency maps, we use a \emph{cross-attention module} to promote alignment between the concept saliency maps $S_\vc$ and image feature summary projected onto the concept subspace. For class-level explanations, we extract the classifier weight $\vw_l$ from the Fully Connected Layer (FCL) for class $l$ and linearly combine the concept saliency maps, \ie $S_l \defn \sum \vw_l S_\vc$.

\paragraph{Loss Function.}
To ensure both concept- and class-level accuracy while maintaining the locality faithfulness of the generated saliency maps, SL-CBM is trained using the following loss functions: 1) \textbf{Class Accuracy}: We use standard \emph{Cross-Entropy Loss} \[\cL_{ce} \defn -g(f(\vx))_{l_{gt}} + \log\sum_j \exp(g(f(\vx)))_j)\] to enforce the class precisions. 2) \textbf{Concept Accuracy}: \emph{Concept Accuracy Loss} $\cL_{ca}$ is defined as \[ \cL_{ca} \defn \cL_1(\gamma (s(f(\vx)) - s(\mathbb{1}(\cC_{gt})))), \] where $s(\cdot)$ is softmax function, $\cL_1(\cdot)$ is standard mean absolute error, $\gamma$ scales the loss, and $\mathbb{1}(\cC_{gt})$ is the indicator vector of ground-truth concepts. 3) \textbf{Saliency Sparisity}: To encourage concise and meaningful explanations, \emph{Entropy Loss} \[\cL_e \defn \sum_{i,j} H(s(S_\vc^{(i,j)})),\] is applied over spatial positions $(i,j)$ of the concept saliency map $S_\vc$, where $H(\cdot)$ demotes the entropy.
4) \textbf{Optional: Concept Consistency}:
To further refine SL-CBM, we optionally include \emph{Contrastive Loss} \[ \cL_c \defn \frac{1}{n}\sum_{i,j}-\log \frac{e^{sim(f_i,f_j)/\tau}}{\sum_{m \neq i} e^{sim(f_i,f_m)/\tau}},\] to promote intra-class similarity and inter-class distinction of concept embeddings. $sim(f_i,f_j)$ represents the similarity between the concpet embeddings $f(\vx)_i$ and $f(\vx)_j$, $n$ is the number of examples (mini-batch), and $\tau$ is the temperature parameter.
The total loss combines these terms, \[\cL \defn \lambda_{ce}\cL_{ce} + \lambda_{ca}\cL_{ca} + \lambda_e\cL_e + \lambda_c\cL_c, \] balancing classification accuracy, concept fidelity, and saliency map interpretability.

\subsection{Evaluating Locality Faithfulness}
To fairly assess locality faithfulness, we use two setups: 1) with ground truth segmentation annotations at both the concept and class levels, and 2) without such annotations.

In the first scenario, saliency maps at both levels are denoted as $S$, and corresponding binary ground truth masks as $M$, where spatial locations contributing to a concept or class are marked $1$, otherwise $0$.
Saliency maps $S$ are binarized as $B \defn \begin{cases}
1 & \text{if } S^{(i,j)} > 0.5, \\
0 & \text{otherwise}.
\end{cases}$ 
Evaluation uses \emph{Intersection over Union} (IoU) and \emph{Dice coefficient}, measuring overlap between  $B$ and $M$. To address inflated scores caused by large saliency regions, we introduce \emph{Compact IoU} (C-IoU), replacing the denominator with the area of $B$ alone. Additionally, all metrics are weighted by image classification accuracy to integrate predictive correctness.

In the second scenario without annotations, saliency maps undergo max-min normalization per standard practice~\cite{zhang2024saliency}. Faithfulness is assessed via Average Drop (AD), Average Increase (AI)\cite{chattopadhay2018grad}, and Average Gain (AG)\cite{zhang2024opti}. AD measures the decrease in class probability when masking the image, AI measures the fraction of cases where masking increases probability, and AG quantifies the overall gain in predictive power. AG is preferred as it reliably detects adversarial cases like FakeCAM~\cite{poppi2021revisiting}, unlike AD and AI, making it a more robust metric for saliency evaluation in explainable AI.

\begin{table*}[bhpt]
    \centering
    \setlength{\tabcolsep}{2pt}
    \begin{tabular}{lr|cc|cc|cc|cc|cc|cc|cc|cc} \toprule
     \multicolumn{2}{c|}{\mr{4}{Method}} & \multicolumn{2}{c|}{\mr{3}{Accuracy}} &\multicolumn{2}{c|}{\mr{3}{Interpretable }} & \mc{12}{Locality Faithfulness} \\\cmidrule{7-18}
     &&\multicolumn{2}{c|}{}&\multicolumn{2}{c|}{\mr{3}{Prediction}} &\multicolumn{6}{c|}{With Annotation}& \mc{6}{Without Annotation}\\\cmidrule{7-18}
     &&\multicolumn{2}{c|}{} &&&\multicolumn{2}{c|}{IoU $\uparrow$}&\multicolumn{2}{c|}{ Dice $\uparrow$}&\multicolumn{2}{c|}{ C-IoU $\uparrow$}&\multicolumn{2}{c|}{ AD $\downarrow$}&\multicolumn{2}{c|}{AI $\uparrow$}&\multicolumn{2}{c}{AG $\uparrow$}\\ \cmidrule{3-18}
     && Concept&Class & NEC-5& ANEC &$S_{\cC_{gt}}$&$S_{l_{gt}}$&$S_{\cC_{gt}}$&$S_{l_{gt}}$&$S_{\cC_{gt}}$&$S_{l_{gt}}$&$S_{\cC_{gt}}$&$S_{l_{gt}}$&$S_{\cC_{gt}}$&$S_{l_{gt}}$&$S_{\cC_{gt}}$&$S_{l_{gt}}$\\\midrule
         \mr{2}{PCBM}&ResNet50 & 49.75 &	89.05 & 51.49 & 71.06 &	6.16 & 0.09 &	9.90 & 0.14 &	21.61 &  0.45 & 0.62 & 0.63 & 46.73 & 46.17 & 0.67 & 0.65 \\
         &ViT-B16 &  39.97 &	90.18 & 18.12 & 54.65 & 4.65& 0.00 &	7.34&  0.00 & 13.87 & 0.15 & \textbf{0.55} &\textbf{0.57}& \textbf{65.38}& \textbf{63.64} & 9.69 &8.94\\\midrule
        \multicolumn{2}{c|}{CSS} & 85.69 & 98.68 & 95.08 & 97.45 &	17.13 &  0.00 & 25.76 &  0.00 & 39.32 & 0.27 &  7.30 & 7.29 & 34.28&  34.42 & 10.53 &10.87\\ \midrule
         \rowcolor{cyan!10}
          \multicolumn{2}{c|}{SL-CBM}& \textbf{90.16} & \textbf{99.11} & \textbf{98.83}& \textbf{99.03} & \textbf{16.21}& \textbf{0.14}	& \textbf{24.44}&\textbf{0.21} & \textbf{44.57} & \textbf{0.47} & 3.65 & 3.68 & 47.87 &  48.07 & \textbf{13.36}& \textbf{13.58}\\
         \bottomrule
    \end{tabular}
    \caption{Comparison of SL-CBM with state-of-the-art CBMs on RIVAL-10 in terms of accuracy, as well as locality faithfulness metrics with annotation (IoU, Dice, C-IoU) and without annotation (AD, AI, AG) at both concept-level and class-level. $\uparrow$ signifies that a higher value is preferable for the metric, while $\downarrow$ indicates that a lower value is better. All values are presented as percentages. The best results are highlighted in bold.}
    \label{tab:cp-Rival10}
\end{table*}
\begin{figure*}[htpb]
    \centering
    \setlength{\tabcolsep}{1.5pt}
    \begin{tabular}{c c c c c c c c c}
    & $\vx$ & $S_{c^1}$ & $S_{c^2}$& $S_{c^3}$ & $S_{c^4}$ & $S_{c^5}$ & $S_l$ \\  
    \rotatebox{90}{PCBM {\scriptsize(ResNet50)}} &
    \includegraphics[width=0.13\linewidth]{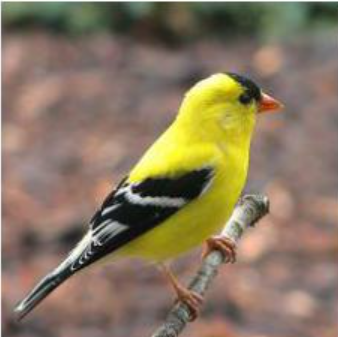}& 
    \includegraphics[width=0.13\linewidth]{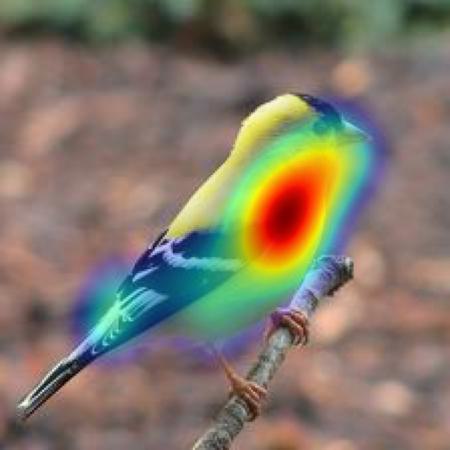}&    
    \includegraphics[width=0.13\linewidth]{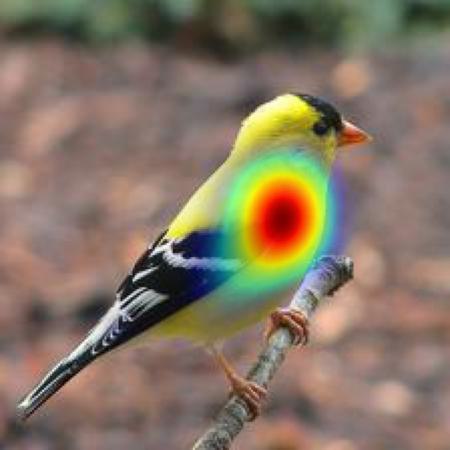}& 
    \includegraphics[width=0.13\linewidth]{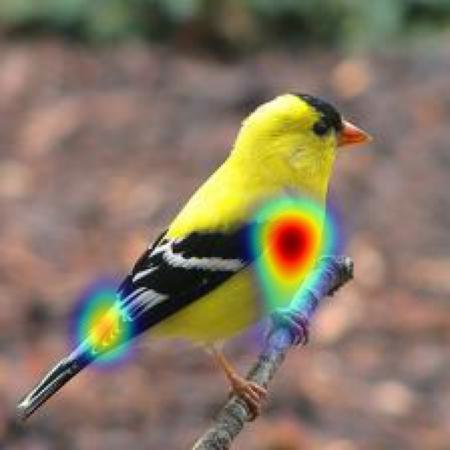}& 
    \includegraphics[width=0.13\linewidth]{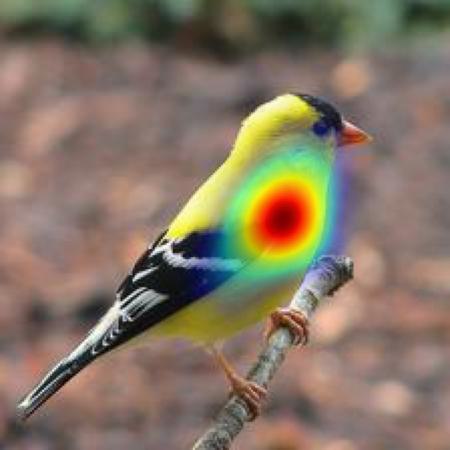}& 
    \includegraphics[width=0.13\linewidth]{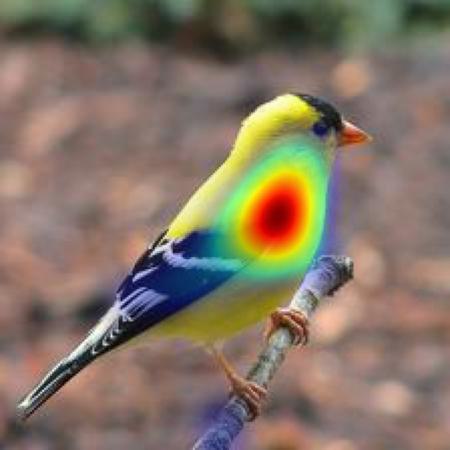}& 
    \includegraphics[width=0.13\linewidth]{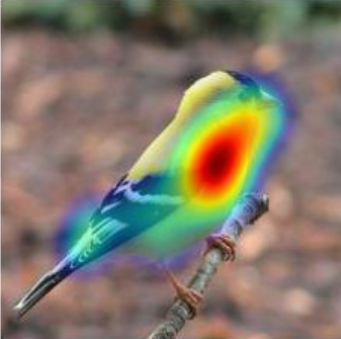}\\
    &  & Beak & Text&  Long &  Colored-eyes &  Hairy &  Bird \\ 
    \rotatebox{90}{PCBM {\scriptsize (ViT-B16)}}  &
    \includegraphics[width=0.13\linewidth]{figures/rival_comparison/original.png}& 
    \includegraphics[width=0.13\linewidth]{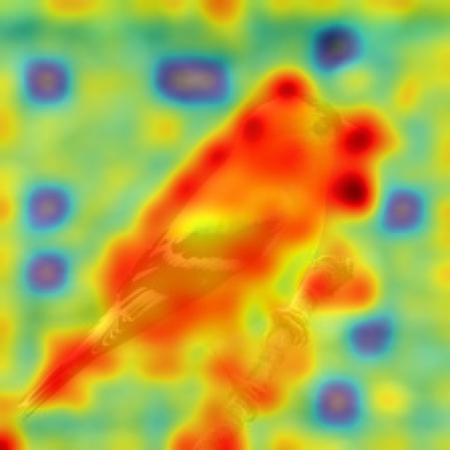}&    
    \includegraphics[width=0.13\linewidth]{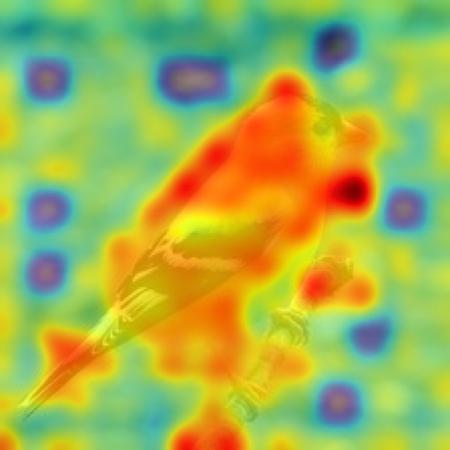}& 
    \includegraphics[width=0.13\linewidth]{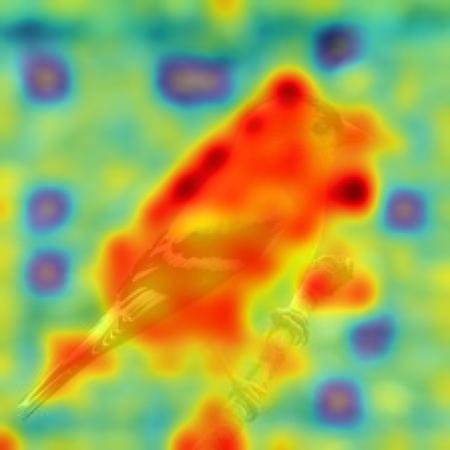}& 
    \includegraphics[width=0.13\linewidth]{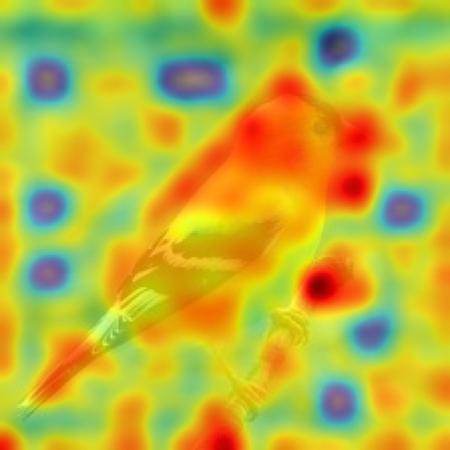}& 
    \includegraphics[width=0.13\linewidth]{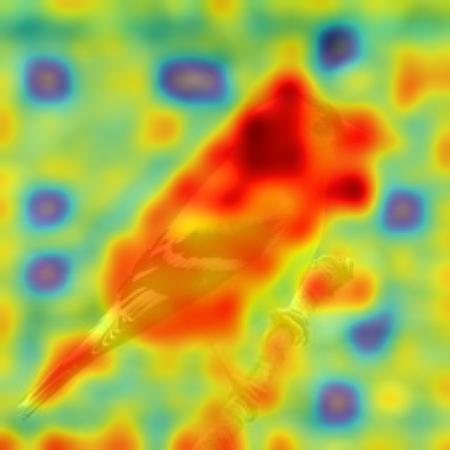}& 
    \includegraphics[width=0.13\linewidth]{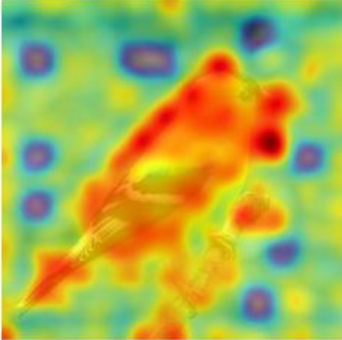}\\
    &  & Beak & Tall&  Long &  Long-snout &  Wet & Bird  \\ 
    \rotatebox{90}{CSS}  &
    \includegraphics[width=0.13\linewidth]{figures/rival_comparison/original.png}& 
    \includegraphics[width=0.13\linewidth]{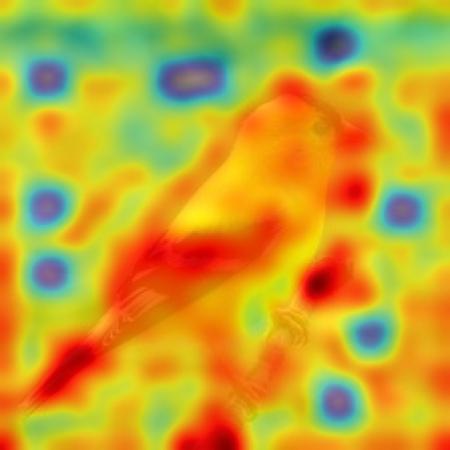}&    
    \includegraphics[width=0.13\linewidth]{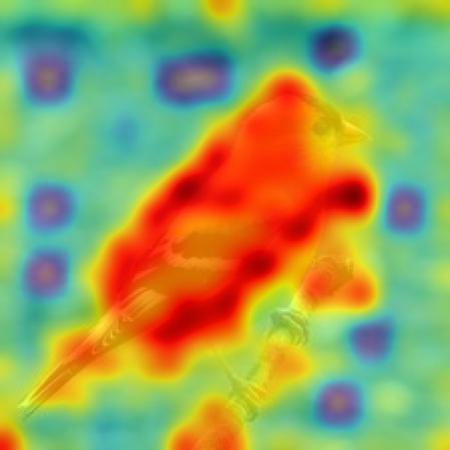}& 
    \includegraphics[width=0.13\linewidth]{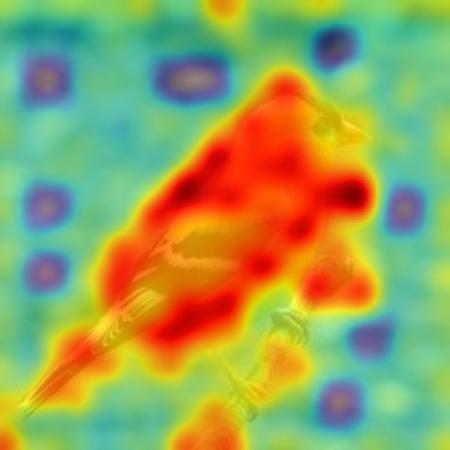}& 
    \includegraphics[width=0.13\linewidth]{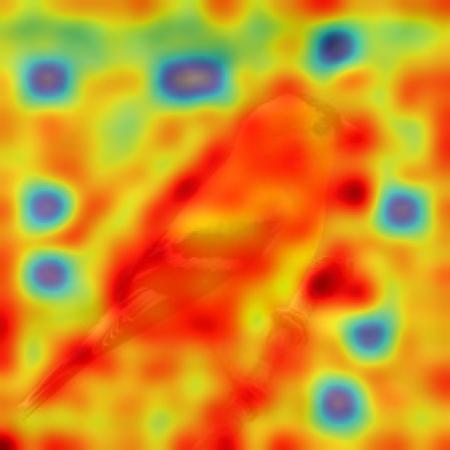}& 
    \includegraphics[width=0.13\linewidth]{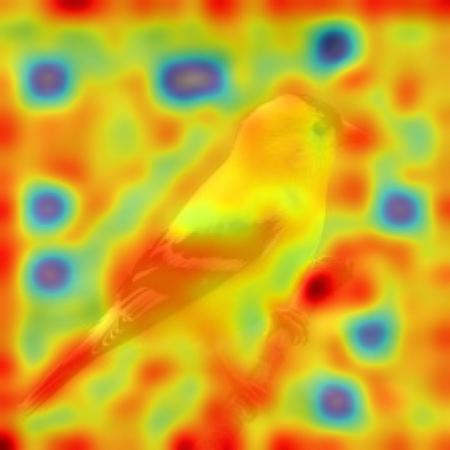}& 
    \includegraphics[width=0.13\linewidth]{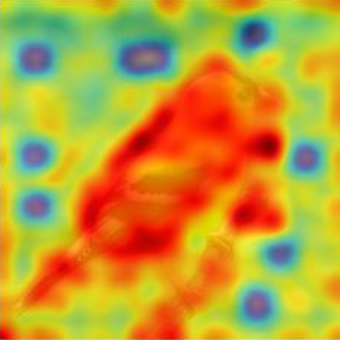}\\
    &  & Patterned & Beak&  Wings &  Tail &  Horns & Bird  \\ 
    \rotatebox{90}{SL-CBM(ours)}  &
    \includegraphics[width=0.13\linewidth]{figures/rival_comparison/original.png}& 
    \includegraphics[width=0.13\linewidth]{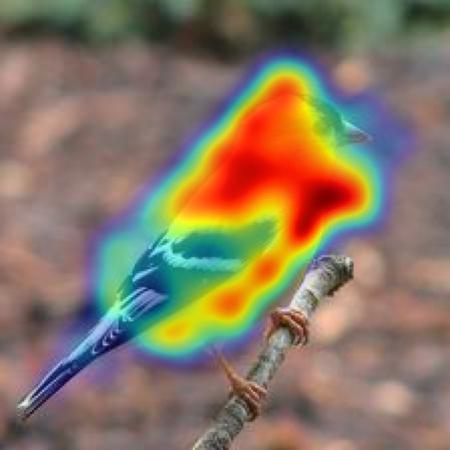}&    
    \includegraphics[width=0.13\linewidth]{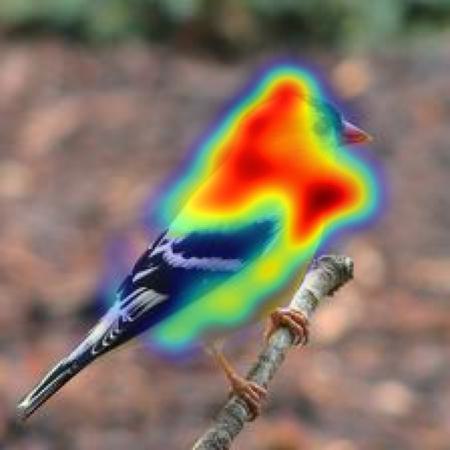}& 
    \includegraphics[width=0.13\linewidth]{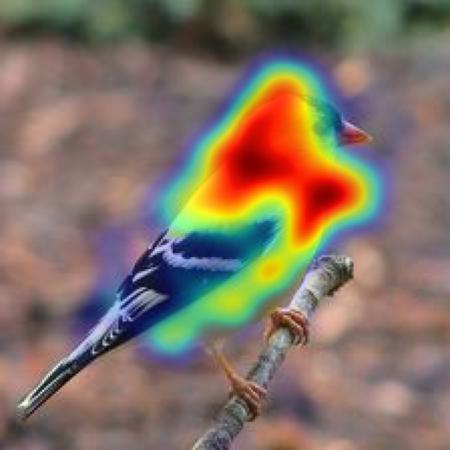}& 
    \includegraphics[width=0.13\linewidth]{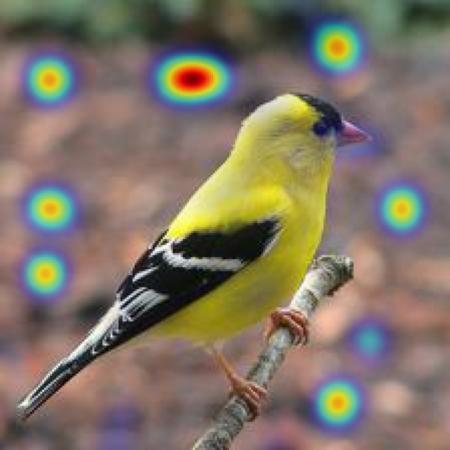}& 
    \includegraphics[width=0.13\linewidth]{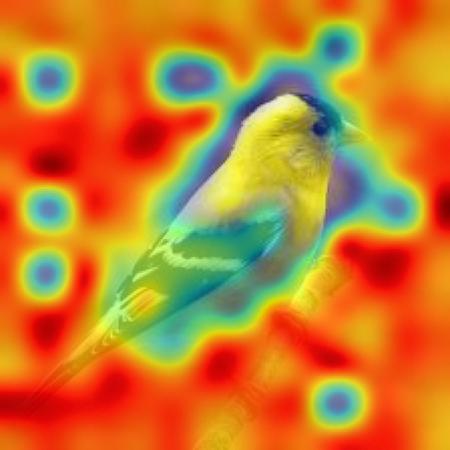}& 
    \includegraphics[width=0.13\linewidth]{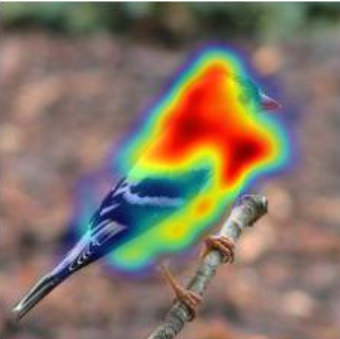}\\
    &  & Beak & Patterned &  Wings &  Tail &  Rectangular & Bird  \\ 
    \end{tabular}    
    \caption{We present the saliency maps on RIVAL-10, showing the saliency maps for the top 5 concepts, where $c^1$ represents the top predicted concept, $c^2$ the second top, and so on, along with the predicted class $l$. The CBM saliency maps are generated using GradCAM~\cite{selvaraju2017grad}, while the SL-CBM saliency maps are produced by our proposed method.}
    \label{fig:top5-concept-class}
\end{figure*}

\section{Experiments}

\paragraph{Dataset.}
We use the RIVAL-10 dataset~\cite{moayeri2022comprehensive}, which maps CIFAR-10~\cite{krizhevsky2009learning} classes to ImageNet~\cite{krizhevsky2012imagenet} and contains around 26,000 images with 18 visual attributes and segmentations. We train on the training dataset (21,098 images) and evaluate on the test set (5,286 images), reporting all results on the test set.
Additionally, we use the CUB-200-2011 dataset~\cite{WahCUB_200_2011}, a fine-grained categorization benchmark with 11,788 images across 200 bird subcategories, annotated with part locations, attributes, and bounding boxes. We train on 4,796 images and test on 5,794, following PCBM~\cite{yuksekgonul2022post} by using an imbalanced dataset sampler.

\paragraph{Models.} 
We compare SL-CBM with two state-of-the-art CBMs: PCBM~\cite{yuksekgonul2022post} and CCS~\cite{selvaraj2024improving}. We evaluate PCBM with ResNet50 and ViT-B16 backbones, while CSS uses a ViT-B16-based CLIP model with its original training parameters.
SL-CBM employs a ViT-B16-based CLIP model and is trained with a learning rate of $0.0003$ using the Adam optimizer. By default, we set $\lambda_{ce}=1$, $\lambda_{ca}=10^4$, $\lambda_e=5$, and $\lambda_c=0$.
To test CBM adaptability, we also use ResNet18\footnote{Pre-trained weights from torchcv~\cite{you2019torchcv}} as a backbone, adjusting SL-CBM's $\lambda_e$ to 1.0 while keeping other parameters unchanged.

\paragraph{Environment Settings.}
Experiments ran on Ubuntu 22.04.5 with a Xeon 8336C CPU, 125GB RAM, 127GB swap, and four RTX 4090 GPUs (24GB VRAM each), using Python 3.12.11, CUDA 12.4, and PyTorch 2.x with GPU acceleration. Unless noted, a single RTX 4090 was used.

\paragraph{Evaluation Protocol.}
Beyond standard concept- and class-level accuracy, we assess interpretable prediction using NEC-5 and ANEC metrics~\cite{srivastava2024vlg}, which measure accuracy based on decision-related concepts: NEC-5 
uses the top 5, while ANEC averages over varying counts in $[5,10,15]$ for on RIVAL-10 and $[5,10,15,20,25,30]$ for CUB.
For locality faithfulness, we compute IoU, Dice coefficient, and C-IoU between saliency maps and segmentation masks, along with AI, AD, and AG metrics without segmentation masks. For CBMs lacking native saliency map outputs, we use Grad-CAM~\cite{selvaraju2017grad} to generate them.
Concept evaluation is restricted to saliency maps of ground truth concepts; class evaluation considers only the ground truth class.

\begin{table}[]
    \centering
    \small
    \setlength{\tabcolsep}{0.2pt}
    \begin{tabular}{l|cc|cc|cc|cc|cc} \toprule
     {\mr{3}{Method}}  & \multicolumn{2}{c|}{\mr{2}{Accuracy}}&\multicolumn{2}{c|}{Interpretable} & \mc{6}{Locality Faithfulness} \\\cmidrule{6-11}
     &  &  &\multicolumn{2}{c|}{Prediction}& \multicolumn{2}{c|}{AD $\downarrow$}& \multicolumn{2}{c|}{AI $\uparrow$}& \mc{2}{AG $\uparrow$}\\\cmidrule{2-11}
     & {\small Concept} & {\small Class}& NEC-5& ANEC & $S_{\cC_{gt}}$&$S_{l_{gt}}$& $S_{\cC_{gt}}$&$S_{l_{gt}}$& $S_{\cC_{gt}}$&$S_{l_{gt}}$ \\\midrule
    PCBM      & {69.4} & 58.9 & 15.1 & 37.1 &7.9&  7.8 & 28.7& 29.0 & 2.5& 2.5 \\\midrule
        CSS  &  59.2 & 51.2  & 17.9 & 34.6 & 6.8 & 5.6& \textbf{57.5}& \textbf{58.8} & 9.00 & 8.3\\ \midrule
         \rowcolor{cyan!10}
     SL-CBM  & \textbf{83.7} & \textbf{60.9} & \textbf{38.0} & \textbf{51.6} & \textbf{4.2} & \textbf{3.2} & 49.9 & 51.1 & \textbf{10.6} & \textbf{9.8}\\
         \bottomrule
    \end{tabular}
    \caption{Comparison of SL-CBM with state-of-the-art CBMs on CUB in terms of accuracy, as well as locality faithfulness metrics without annotation (AD, AI, AG) at both concept-level and class-level. $\uparrow$ signifies that a higher value is preferable for the metric, while $\downarrow$ indicates that a lower value is better. All values are presented as percentages. The best results are highlighted in bold.}
    \label{tab:cp-CUB}
\end{table}

\subsection{Comparison}
We compare SL-CBM with state-of-the-art CBMs on RIVAL-10, as it provides segmentation annotations at both the concept and class levels. This allows us to evaluate accuracy at both levels, along with locality faithfulness, using annotated and non-annotated metrics. The results are presented in Table \ref{tab:cp-Rival10}. SL-CBM achieves the highest class- and concept-level accuracy as well as interpretable prediction, \ie NEC-5 and ANEC. With segmentation annotations, SL-CBM surpasses other methods in locality faithfulness. In the absence of accurate annotations, we use AD, AI, and AG as surrogate metrics, where SL-CBM performs well overall, particularly excelling in AG, the most reliable measures of locality faithfulness without annotation. 

SL-CBM effectively leverages pre-trained backbones. 
On CUB with a pre-trained ResNet18, we compare SL-CBM with other state-of-the-art CBMs using the same backbone.
PCBM achieves strong and consistent performance in both accuracy and locality faithfulness in Table \ref{tab:cp-CUB} compared to Table \ref{tab:cp-Rival10}, while CSS shows degraded performance, with the lowest concept and class accuracy. SL-CBM outperforms others except in AI, which is less reliable than AG despite similar principles. Its superior locality faithfulness, especially in AG, highlights robustness across pre-trained backbones.

\paragraph{Visualization.} 
Figure \ref{fig:top5-concept-class} presents saliency maps at both concept and class levels, highlighting the top five concepts per CBM. 
Compared to CNN and transformer-based PCBM models, CNN backbones produce more focused saliency maps. PCBM with ViT-B16, sharing the same backbone as CSS and SL-CBM, shows that GradCAM produces unreliable maps for both PCBM and CSS, lacking proper localization despite correct predictions. In contrast, SL-CBM improves both localization and prediction. For example, its saliency map for the concept \emph{Break} identifies relevant regions, unlike other transformer-based CBMs, which often highlight the entire image. SL-CBM also produces more distinct concept maps than PCBM and CSS.
While not perfect, SL-CBM inherently generates concept saliency maps, enabling failure diagnosis, \eg, failing to learn \emph{Tail} in this specific image. This diagnosis is unclear in other CBMs due to reliance on GradCAM. 

\begin{figure}[htpb]
    \centering
    \setlength{\tabcolsep}{1pt}
    \renewcommand{\arraystretch}{0}
    \begin{tabular}{lr}
    
    \begin{tikzpicture}
    \begin{axis}[width=.5\columnwidth, height=3.1cm,
    xlabel=$\lambda_e$,
    ylabel=concept accuracy,
    xmode=log,
    legend columns=2,
    legend pos=outer north east]
    \addplot[mark=star,black]
        table [x=E, y=concept_accuracy, col sep=space]{figures/ablation-E.txt};  
    \addplot[mark=o,blue]
        table [x=E, y=concept_accuracy, col sep=space]{figures/ablation-E-C.txt};
    \end{axis}
    \end{tikzpicture}&

     \begin{tikzpicture}
    \begin{axis}[width=.5\columnwidth, height=3.1cm,
    xlabel=$\lambda_e$,
    xmode=log,
    ylabel=class accuracy,]
    \addplot[mark=star,black]
        table [x=E, y=class_accuracy, col sep=space]{figures/ablation-E.txt}; 
    \addplot[ mark=o,blue]
        table [x=E, y=class_accuracy, col sep=space]{figures/ablation-E-C.txt};
    \end{axis}
    \end{tikzpicture} \\ 

      \begin{tikzpicture}
    \begin{axis}[width=.5\columnwidth, height=3.1cm,
    xlabel=$\lambda_e$,
    ylabel=IoU $\uparrow$,
    xmode=log,
    legend columns=2,
    legend pos=outer north east]
    \addplot[mark=star,black]
        table [x=E, y=IoU, col sep=space]{figures/ablation-E.txt};
    \addplot[mark=o,blue]
        table [x=E, y=IoU, col sep=space]{figures/ablation-E-C.txt};  
    \end{axis}
    \end{tikzpicture}&

    \begin{tikzpicture}
    \begin{axis}[width=.5\columnwidth, height=3.1cm,
    xlabel=$\lambda_e$,
    xmode=log,
    ylabel=Dice $\uparrow$,
    legend columns=2,
    legend pos=outer north east]
    \addplot[mark=star,black]
        table [x=E, y=Dice, col sep=space]{figures/ablation-E.txt}; 
    \addplot[mark=o,blue]
        table [x=E, y=Dice, col sep=space]{figures/ablation-E-C.txt}; 
    \end{axis}
    \end{tikzpicture}\\

    \begin{tikzpicture}
    \begin{axis}[width=.5\columnwidth, height=3.1cm,
    xlabel=$\lambda_e$,
    xmode=log,
    ylabel=C-IoU $\uparrow$,]
    \addplot[mark=star,black]
        table [x=E, y=C-IoU, col sep=space]{figures/ablation-E.txt};
    \addplot[mark=o,blue]
        table [x=E, y=C-IoU, col sep=space]{figures/ablation-E-C.txt};
    \end{axis}
    \end{tikzpicture}

    &
     \begin{tikzpicture}
    \begin{axis}[width=.5\columnwidth, height=3.1cm,
    xlabel=$\lambda_e$,
    ylabel=AD $\downarrow$,
    xmode=log,
    legend columns=2,
    legend pos=outer north east]
    \addplot[mark=star,black]
        table [x=E, y=AD, col sep=space]{figures/ablation-E.txt}; 
    \addplot[mark=o,blue]
        table [x=E, y=AD, col sep=space]{figures/ablation-E-C.txt};   
    \end{axis}
    \end{tikzpicture} \\

     \begin{tikzpicture}
    \begin{axis}[width=.5\columnwidth, height=3.1cm,
    xlabel=$\lambda_e$,
    ylabel=AI $\uparrow$,
    xmode=log,
    legend columns=2,
    legend pos=outer north east]
    \addplot[mark=star,black]
        table [x=E, y=AI, col sep=space]{figures/ablation-E.txt};
    \addplot[mark=o,blue]
        table [x=E, y=AI, col sep=space]{figures/ablation-E-C.txt};  
    \end{axis}
    \end{tikzpicture}

    & \begin{tikzpicture}
    \begin{axis}[width=.5\columnwidth, height=3.1cm,
    xlabel=$\lambda_e$,
    ylabel=AG $\uparrow$,
    xmode=log,
    legend columns=2,
    legend pos=outer north east]
    \addplot[mark=star,black]
        table [x=E, y=AG, col sep=space]{figures/ablation-E.txt};
    \addplot[mark=o,blue]
        table [x=E, y=AG, col sep=space]{figures/ablation-E-C.txt};  
    \end{axis}
    \end{tikzpicture} \\
    \multicolumn{2}{c}{
\begin{tikzpicture}
\begin{axis}[
    hide axis,
    xmin=0, xmax=1,
    ymin=0, ymax=1,
    legend columns=3,
    legend style={at={(0.8,-0.2)}, anchor=north},
    legend entries={$\lambda_c=0$ ,$\lambda_c=1$ }
]

    \addlegendimage{mark=star}
    \addlegendimage{color=blue,mark=o}
\end{axis}
\end{tikzpicture}
}
    \end{tabular}
\caption{Ablation study on $\lambda_e$ analyzing its effect on class accuracy, concept accuracy, IoU, Dice, and C-IoU with/without $\cL_c$ ($\lambda_c = 1$ or $0$). Experiments are on RIVAL-10 with $\lambda_{ce} = 1$ and $\lambda_{ca} = 10^4$. $\uparrow$ signifies that a higher value is preferable for the metric, while $\downarrow$ indicates that a lower value is better. All values are presented as percentages.}
\label{fig:ablation-E}
\end{figure}
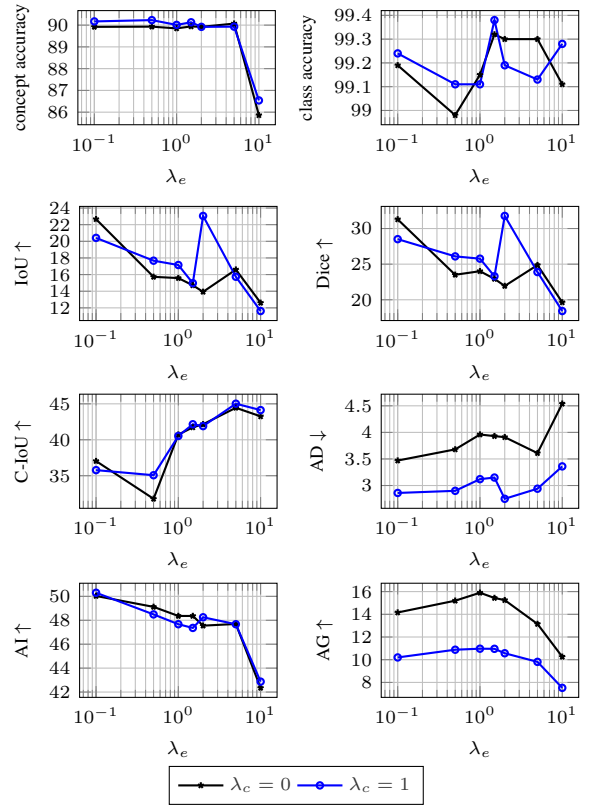

\begin{figure}[htpb]
    \centering
    \setlength{\tabcolsep}{1pt}
    \begin{tabular}{c c c}
    Original Image & $\lambda_c=0$  & $\lambda_c=1$ \\  
    \includegraphics[width=0.26\linewidth]{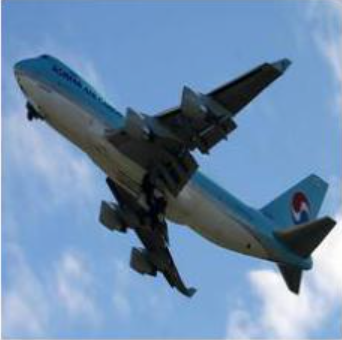}& 
    \includegraphics[width=0.26\linewidth]{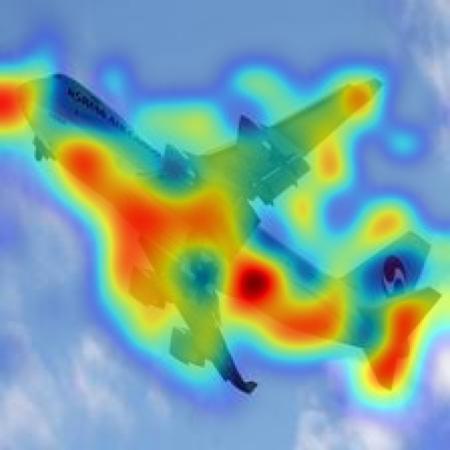}&   
    \includegraphics[width=0.26\linewidth]{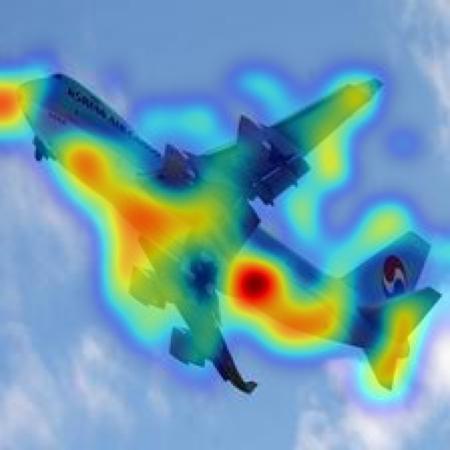} 
    \end{tabular}    
    \caption{We present a top 1 concept saliency map on RIVAL-10 of $\lambda_c = 0 \text{ or } 1$.}
    \label{fig:c-loss}
\end{figure}

\subsection{Ablation}
We perform an ablation study on $\lambda_e$ and $\lambda_c$, while keeping $\lambda_{ce} = 1$ for Cross-Entropy Loss and $\lambda_{ca} = 10^4$ for Concept Accuracy Loss, following the optimal settings of CCS~\cite{selvaraj2024improving}. Specifically, we vary $\lambda_e$ for Entropy Loss over the set $\{0, 0.1, 0.5, 1.0, 1.5, 2.0, 5.0, 10.0\}$ and set $\lambda_c$ for Contrastive Loss to either $1$ or $0$.
We evaluate these configurations on RIVAL-10, focusing on concept-level metrics as concept and class locality faithfulness show similar trends. The results, depicted in Figure \ref{fig:ablation-E}, reveal that including Contrastive Loss ($\lambda_c = 1$) leads to more unstable performance in class accuracy, IoU, and Dice scores, while maintaining comparable concept accuracy, C-IoU, and AI. Without Contrastive Loss ($\lambda_c = 0$), the model shows improved AG but reduced AD.

Visualization in Figure \ref{fig:c-loss} further illustrates that when $\lambda_c = 0$, saliency maps highlight larger image regions, making AD and AG metrics more sensitive to these differences. Additionally, excessively large $\lambda_e$ values cause significant performance degradation by producing overly sparse saliency maps that hinder sufficient learning. Balancing the trade-offs across all metrics,, we identify a local optimum at $\lambda_e = 5.0$.
To demonstrate the adaptability of our method to new datasets and models, we conduct an experiment showing that parameters can be efficiently optimized using a small dataset and then successfully applied to the full dataset. Due to space constraints, these results are presented in the supplementary material.

\begin{figure}[tbph]
    \centering
    \setlength{\tabcolsep}{0pt}
    \begin{tabular}{cc}
      \begin{tikzpicture}
\begin{axis}[width=.5\columnwidth, height=3cm,
    xlabel={Intervention Counts},
    ylabel={Task Error},
    grid=major,
    legend pos=north west,
]


\interventionPlot{figures/intervention/PCBM_RN18/CUB/ag}{color=red}{ag}
\interventionPlot{figures/intervention/PCBM_RN18/CUB/cctp}{color=black}{cctp}
\interventionPlot{figures/intervention/PCBM_RN18/CUB/lcp}{color=yellow}{lcp}
\interventionPlot{figures/intervention/PCBM_RN18/CUB/rand}{color=green}{rand}
\interventionPlot{figures/intervention/PCBM_RN18/CUB/ucp}{color=brown}{ucp}
\legend{}

\end{axis}
\end{tikzpicture}   &
\begin{tikzpicture}
\begin{axis}[width=.5\columnwidth, height=3cm,
    xlabel={Intervention Counts},
    ylabel={Task Error},
    grid=major,
    legend pos=north west,
]


\interventionPlot{figures/intervention/CSS/CUB/ag}{color=red}{ag}
\interventionPlot{figures/intervention/CSS/CUB/cctp}{color=black}{cctp}
\interventionPlot{figures/intervention/CSS/CUB/lcp}{color=yellow}{lcp}
\interventionPlot{figures/intervention/CSS/CUB/rand}{color=green}{rand}
\interventionPlot{figures/intervention/CSS/CUB/ucp}{color=brown}{ucp}
\legend{}

\end{axis}
\end{tikzpicture}
         \\
         PCBM &
CSS
         \\
\mc{2}{
\begin{tikzpicture}
\begin{axis}[width=.5\columnwidth, height=3cm,
    xlabel={Intervention Counts},
    ylabel={Task Error},
    grid=major,
    legend style={
        at={(1.05,0.5)}, 
        anchor=west      
    },
    legend cell align=left,
]


\interventionPlot{figures/intervention/SPSS/CUB/ag}{color=red}{ag}
\interventionPlot{figures/intervention/SPSS/CUB/cctp}{color=black}{cctp}
\interventionPlot{figures/intervention/SPSS/CUB/lcp}{color=yellow}{lcp}
\interventionPlot{figures/intervention/SPSS/CUB/rand}{color=green}{rand}
\interventionPlot{figures/intervention/SPSS/CUB/ucp}{color=brown}{ucp}

\end{axis}
\end{tikzpicture} 
\begin{adjustbox}{raise=1cm}
\begin{tikzpicture}
\begin{axis}[
    hide axis,
    legend columns=2,
    xmin=0, xmax=120,
    ymin=0, ymax=1.05,
    ytick distance=0.2,
    legend style={at={(0,1)}, anchor=north west}
]

\addlegendimage{color=green}
\addlegendentry{RAND}
\addlegendimage{color=brown}
\addlegendentry{UCP}
\addlegendimage{color=yellow}
\addlegendentry{LCP}
\addlegendimage{color=black}
\addlegendentry{CCTP}
\addlegendimage{color=red}
\addlegendentry{AG}

\end{axis}
\end{tikzpicture}
\end{adjustbox}}
\\
         \mc{2}{SL-CBM}
    \end{tabular}
    \caption{Intervention effectiveness on CUB: greater error reduction for the same number of concepts corrected.}
    \label{fig:intervention}
\end{figure}
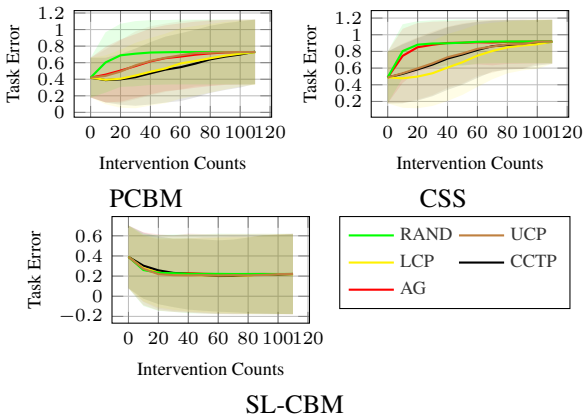

\subsection{Intervention}
Intervention plays a critical role in correcting model outputs externally, thereby enhancing the model's applicability in real-world scenarios. Following the approach of \citet{shin2023closer}, we perform interventions by replacing predicted concepts with their ground truth values. We explore both random intervention (RAND) and guided interventions, where concepts are ranked using various metrics: Uncertainty in Concept Predictions (UCP), Loss on Concept Prediction (LCP), and Contribution of Concept to Target Prediction (CCTP). Additionally, we test the use of the explainability metric, \ie AG, to rank and replace the top concepts with ground truth values. These experiments are conducted on the CUB dataset, which offers a sufficient number of concepts for meaningful intervention analysis. As shown in Figure~\ref{fig:intervention}, AG shows limited effectiveness as a metric for guiding intervention. Overall, only our proposed SL-CBM benefits from intervention, while PCBM and CSS experience degraded performance. When a model fails to learn concepts faithfully, intervention can be detrimental. In contrast, by enforcing local faithfulness, SL-CBM is better aligned with concept learning, making intervention more effective. 
For intervention counts, we test values in $[0, 10, 20, 30, 40, 50, 60, 70, 80, 90, 100, 110]$ on CUB.

\section{Conclusion}

In this work, we present SL-CBM, a novel extension of CBMs that significantly enhances locality faithfulness by generating both concept-level and class-level saliency maps. While traditional CBMs offer concept-based explanations, they often lack meaningful spatial alignment between concepts and relevant image regions. SL-CBM overcomes this limitation through the integration of a $1 \times 1$ convolution and a cross-attention mechanism, which together improve spatial coherence and model interpretability.
By providing saliency maps that faithfully reflect the model’s internal reasoning, SL-CBM enables effective debugging: poor saliency quality signals concept learning failure, addressing a critical shortcoming of post-hoc explanation methods. Our experiments show that SL-CBM outperforms state-of-the-art CBMs in accuracy, locality faithfulness, and intervention performance, confirming that enforcing locality faithfulness improves concept faithfulness.

Moreover, our ablation studies reveal the essential role of contrastive loss and entropy regularization in balancing prediction accuracy, explanation faithfulness, and sparsity of saliency maps. Nevertheless, SL-CBM’s effectiveness remains contingent on the quality of the predefined concept set, as it currently does not incorporate concept refinement strategies. Future work could explore integrating concept discovery and refinement to enhance explanation quality, or extend the approach toward applications in model oversight.
Overall, SL-CBM represents a significant advancement in explainable AI by bridging concept-based reasoning with spatially-aware, faithful visual explanations. We believe this work lays a robust foundation for future exploration of trustworthy and interpretable concept-based models.


\section{Acknowledgements}
This work is supported by CAS Project for Young Scientists in Basic Research Grant YSBR-040, ISCAS New Cultivation Project ISCAS-PYFX-202201, ISCAS Basic Research ISCAS-JCZD-202302, JST CREST JPMJCR21D3, and JSPS Grand-in-aid 23H00483. This work also received support from DFG under grant No.~389792660 as part of TRR~248\footnote{CPEC:\url{https://perspicuous-computing.science}}, and funding from the European Union’s Horizon 2020 research and innovation programme under the Marie Skłodowska-Curie grant agreement No 1010082337\footnote{MISSION:\url{https://mission-project.eu/}}.

{
    \bibliography{reference}
}

\end{document}